# Treating 'Free Word Order' in Machine Translation

Ralf STEINBERGER

UMIST - Centre for Computational Linguistics
Manchester, UK, ralf@ccl.umist.ac.uk

## 0  Abstract

In *free word order* languages, every sentence is embedded in its specific context. The order of constituents is determined by the categories *theme*, *rheme* and *contrastive focus*. This paper shows how to recognise and to translate these categories automatically on a sentential basis, so that sentence embedding can be achieved without having to refer to the context. Traditionally neglected modifier classes are fully covered by the proposed method.

## 1  Introduction

Most languages known as *free word order languages* are in fact languages with *partially free word order* (Engelkamp et al. 1992), or rather *free phrase order* (Schäufele 1991). A difficulty linked to the formal description of these languages is that instead of a complete lack of ordering rules many subtle restrictions apply. A large amount of word order variations are grammatical in isolated sentences, but context restricts the number of sequences which are possible and natural. In this sense, sentences are *embedded* in their context. A specific context calls for a certain word order, and the word order of a given sentence reflects its context.

In this paper, we present recent suggestions on how to treat free phrase order in Natural Language Processing (NLP), and present an alternative solution to the problem. The idea is to use a *thematically-tagged*, or *flexible*, canonical form (CF) for generation, and an algorithm to recognise the relevant categories *theme*, *rheme* and *contrastive focus* during analysis. This method has been implemented successfully in the unification and constraint-based Machine Translation system CAT2 (Sharp 1989, Steinberger 1992a). It includes the ordering of modifiers, which are traditionally left out in word order description (Conlon/Evens 1992). All statements in this paper concern written language, as spoken language is more liberal with respect to ordering.

## 2  The Data

We shall start by presenting some data which illustrates the problems related to word order treatment in NLP. Many ordering variations are possible (1a - 1e, 2a, 2b), but some of them are less natural (1e), and others are even ungrammatical (2c, 2d). 1e is only acceptable if the personal pronoun *ich* is heavily stressed, indicated here in capitals.[1]

1a  Morgen$_{26}$ werde ich ihn vielleicht$_{12}$ besuchen.
*Tomorrow will I him probably visit*

1b  Ich werde ihn vielleicht$_{12}$ morgen$_{26}$ besuchen.
*I will him probably tomorrow visit*

1c  Ich werde ihn morgen$_{26}$ vielleicht$_{12}$ besuchen.
*I will him tomorrow probably visit*

1d  Vielleicht$_{12}$ werde ich ihn morgen$_{26}$ besuchen.
*Probably will I him tomorrow visit*

1e  ? Morgen$_{22}$ werde ihn vielleicht$_{12}$ ICH besuchen.
*Tomorrow will him probably I visit*

2a  Er fuhr dennoch$_{20}$ ebenfalls$_{35}$ nach München.
*He drove nevertheless also to Munich*

---

[1] The use of the index numbers will be explained in section 5.

2b Dennoch$_{20}$ fuhr er ebenfalls$_{35}$ nach München.
*Nevertheless drove he also to Munich*

2c * Er fuhr ebenfalls$_{35}$ dennoch$_{20}$ nach München.
*He drove also nevertheless to Munich*

2d * Ebenfalls$_{35}$ fuhr er dennoch$_{20}$ nach München.
*Also drove he nevertheless to Munich*

Depending on the context, different word orders are either required or, at the very least, they are more natural than others. Although in 3 and 4 the context is represented by questions, it is not normally limited to these. 3a, which is the most natural answer to 3, is very unnatural, if not ungrammatical, in 4. Although not all contexts restrict the order of constituents as drastically as 3 and 4, it is a general rule for German and similar languages that sentences are more natural if they are properly embedded in their contexts:

3 Wen erwartete die Frau mit dem Nudelholz?
*Whom waited-for the woman with the rolling pin*

3a Die Frau erwartete mit dem Nudelholz ihren MANN.
*The woman waited-for with the rolling pin her husband*

3b ? Die Frau erwartete ihren MANN mit dem Nudelholz.
*The woman waited-for her husband with the rolling pin*

4 Mit was erwartete die Frau ihren Mann?
*With-what waited-for the woman her husband*

4a Die Frau erwartete ihren Mann mit dem NUdelholz.
*The woman waited-for her husband with the rolling pin*

4b ?? Die Frau erwartete mit dem NUdelholz ihren Mann.
*The woman waited-for with the rolling pin her husband*

It is generally acknowledged that the combination of several factors determines the order of constituents in German and similar languages. In Steinberger (1994), eleven principles acting on the pragmatic, semantic and syntactic levels are listed, each of which can be reformulated as one or several linear precedence (LP) rules. The factors comprise of the tendencies to order elements according to the theme-rheme structure and/or to the functional sentence perspective. Furthermore, they concern verb bonding, animacy, heaviness, the importance of semantic roles for phrase ordering, and others. A distinct feature of the ordering regularities is that none of the factors can be formulated as an absolute LP rule, which makes word order description difficult to deal with in NLP. In recent years several propositions were made to deal with this phenomenon in either analysis or generation, or both.

## 3 Recent Suggestions on Treating Free Phrase Order

Uszkoreit (1987) suggests overcoming the lack of absolute rules by using disjunctions of LP rules. The idea is that if at least one LP rule sanctions a sequence of constituents, the sentence is grammatical. The model thus expresses competence, rather than performance, as it either accepts or rejects a sentence, without making a judgement on acceptability differences as in 1.

Another idea put forward by Erbach (1993) accounts for grades of acceptability. Erbach assumes that the order of verb complements ideally is according to an obliqueness hierarchy, and that each deviation from this order decreases the acceptability of the sentence by a factor of 0.8. Two divergences result in an acceptability score of 0.64 (0.8 * 0.8), etc. Problems we see linked to this approach are the use of the obliqueness hierarchy, which limits the preference mechanism to complements, and the fact that every diversion decreases the score invariably, without considering the varying effect of different variations.

A proposal which takes into account the different importance, or weight, of preference rules, is presented in Jacobs (1988). Jacobs assigns each of his preference rules a specific numerical weight. If a rule applies in a given sentence, its value is added to the acceptability score of the sentence, if it is violated, its value is subtracted. The higher the final score, the more natural, or the 'better' the sentence is.

Ideally, all competing preference rules are satisfied. The complication we see with this approach is that some strictly ordered sequences interfere with the calculation of acceptability. Some of them concern the ordering of toners (Abtönungspartikeln; Thurmair 1989) and other modifier subgroups (Steinberger 1994).

Some of the criticism could be overcome by changing the different propositions slightly. For instance, Erbach's (1993) suggestion to add preference to feature-based formalisms could be combined with Uszkoreit's preference rules. An idea to solve the problems linked to Jacobs' weighing mechanism would be to combine it with absolute LP rules, in order to avoid ungrammatical sequences. However, we want to suggest another method, based on our findings concerning natural, marked and ungrammatical word order, and making use of the categories theme, rheme and contrastive focus (henceforth simply called *focus*).

## 4 The New Model

In our approach (cf. Steinberger 1994), we have different ways of dealing with free phrase order in analysis and generation. In analysis (cf. section 6), grammars have to allow most orderings, as barely any phrase order can be completely excluded. Once a structure is assigned to an input sentence, we suggest that thematic, rhematic and contrastively focussed elements be identified by using our insights concerning the recognition of these categories. This information concerning functional sentence perspective can and should be conveyed in the target language of the translation.

With respect to generation (cf. section 5), acceptable orderings are defined by a single comprehensive linear precedence (LP) rule which not only assigns strict priorities to symbols tagged for syntactic category (e.g. N for nominative NP, SIT for situative complement, M for modifier), but also for the thematic categories *theme*, *rheme* and *contrastive focus*. It is crucial that the relative ordering of *syntactic* symbols can be varied by varying their respective *thematic* markings. The LP rule also assigns priorities to syntactic categories which are not thematically marked. Thus, a syntactic element is assigned a default position if no thematic information is available, but is moved out of this default position if thematic information is present. In this way, a single rule represents a fixed canonical form for unmarked elements and at the same time permits widely varying (though not truly *free*) orderings for thematically marked cases.

Generation and analysis according to this method will be presented in more detail now.

## 5 Generation

We argue in Steinberger (1994) that the use of a comprehensive LP rule, as presented in the previous section, is an efficient way of generating sentences which not only are correct in some contexts but which comply with their contextual restrictions. This flexible output is achieved by using the three thematic categories *theme*, *rheme* and *contrastive focus*, which can capture complements as well as modifiers realised by all phrasal categories. Table 1 shows such a CF for German.

The table is to be read from left to right and from top to bottom. The letters N, A, D, G represent the four cases nominative, accusative, dative, and genitive. PO stands for prepositional object, and SIT, DIR and EXP for situative, directional and expansive complements. Nom and Adj are nominal and adjectival complements, M represents the diverse groups of modifiers. The feature +/-d refers to definiteness, +/-a to animacy, SVC to support verb constructions, and the index numbers to M indicate the relative order of modifiers ($M_1$ precedes $M_2$, and so on). The index numbers are based on Hoberg's classification (1981). If elements cannot cooccur, they are separated by a slash (/), as opposed to by an arrow (<).

The CF imposes linear order on an unordered set of arguments and modifiers. When the analysis of the source language fails to recognise theme, rheme and focus, a default order is generated. Although no CF sequence can produce good sentences in all contexts (cf. 3 and 4), the default order is suitable in a large amount of contexts.

$$N_{pron}/N_{+d+b} < (A<D/Nom/Adj)_{pron} < \text{THEME} < N_{+d-a}/N_{-d+a} <$$
$$< (N_{pron}/N_{+d+a})_{+focus}/(A<D)_{pron+focus} < (A<D)_{+d+a} < G_{pron} < N_{-d-a} < (A<D)_{+d-a} <$$
$$< M_{pragm(a1-18)} < M_{sit(a19-40)} < M_{neg(41)} < M_{mod(42-43)}) <$$
$$< (N_{+d-a}/N_{-d}/(A<D)_{+d}/G_{pron}/M_{(a1-43)})_{+rheme} < (A<D)_{-d+a} < a_{mod(44)} <$$
$$< PO_{pron} < (A<D)_{-d-a} < PO_{+d+a} < PO_{+d-a} < PO_{-d+a} < PO_{-d-a} < G_{nom} <$$
$$< (A/D/G/PO/N_{+d-a}/N_{-d}/M_{pragm}/M_{sit}/M_{mod})_{+focus} <$$
$$< \text{SIT/DIR/EXP} < (Nom/Adj)_{-pron} < (N/A/D/G/PO)_{SVC}$$

Table 1: 'Thematically-tagged' Canonical Form for German

Before showing some example sentences generated by this CF, we have to mention one particularity of German, which is that the verb is in second position in declarative matrix clauses (verb-second, or V2 position), and in final position in subordinate clauses (verb-final, or VF position). Nearly any element can take the one position preceding the verb in V2, called the *Vorfeld* ("pre-(verbal) field"). Normally a *thematic* element is placed into the Vorfeld. According to Hoberg's (1981) analysis of the *Mannheimer Duden Korpus*, in 63% of all V2 sentences the nominative complement (subject) takes this place. A convenient way of seeing it is that all elements follow the verb in V2 position according to the CF, and that one (thematic) element is *moved* into the *Vorfeld* position. We suggest that if the analysis of the source language fails to recognise the theme of the sentence, the subject takes this place.

In our model, most elements can either be thematic, rhematic, or *neutral* (i.e. unmarked with respect to theme and rheme). Sentence variations as different as shown in the examples 5a to 5d can be generated using the canonical form presented above, depending on the parameterisation of the features theme, rheme and focus for the different constituents. The order of elements in 5a corresponds to the default order. However, the same order would be generated if the personal pronoun was marked as being thematic, and/or if the adverb *gestern* was rhematic. We put the information +theme in 5a to 5c in brackets to indicate that this feature is not a requirement to generate the respective word orders. The relative order of the adverb and the accusative NP in 5b differs from the one in 5a, because the object *den Mann* is rhematic. In 5c and 5d, *gestern* and *den Mann* are thematic, respectively. In addition to this, the personal pronoun in 5d is marked as being stressed contrastively. We used capital letters to express the obligatory focus. It is easy to think of more phrase order combinations caused by further parameterisations.

5a Ich$_{(+theme)}$ habe den Mann gestern$_{26(+rheme)}$ gesehen. ($A_{+d+a}$ -$M_{26}$)
*I have the man yesterday seen*

5b Ich$_{(+theme)}$ habe gestern$_{26}$ den Mann$_{+rheme}$ gesehen.
*I have yesterday the man seen*

5c Gestern$_{26+theme}$ habe ich den Mann$_{(+rheme)}$ gesehen.
*Yesterday have I the man seen*

5d ..., weil gestern$_{26+theme}$ ICH$_{+focus}$ den Mann gesehen habe.
*... because yesterday I the man seen have.*

Modifiers should be classified according to Hoberg's (1981) 44 modifier position classes, which partly coincide with the common semantic classifications, and partly not. Hoberg's modifier indexes are the result of the statistical verification of Engel's intuitive classes (1970). As modifiers do not always follow in the same order, Hoberg chose a classification which lead to least deviations between her classification and the order in the corpus used (Mannheimer Duden Korpus). The following sentences exemplify the order of the CF for modifiers:

6a Ich habe deshalb$_{22}$ gestern$_{26}$ mit Wolf$_{42}$ ferngesehen.
*I have therefore yesterday with Wolf watched-tv*

6b Ich habe deshalb$_{22}$ mit Wolf$_{42}$ gestern$_{26+rheme}$ ferngesehen.
*I have therefore with Wolf yesterday watched-tv*

7 Damals$_{26+theme}$ bin ich Frauen ohnehin$_9$ oft$_{37}$ überstürzt$_{43}$ davongelaufen.
*Then am I women anyway often overhastyly ranaway (Then, I often ran away from women overhastily anyway)*

Due to the procedure described in this section, ungrammatical sentences such as 2c and 2d can be avoided successfully.

# 6 Analysis

The generation of contextually embedded sentences is based on the successful analysis of theme and rheme constituents. The recognition of contrastive stress is even more important. A basic fact that can be used for the automatic recognition of these categories is that not only the context determines the ordering of constituents in an embedded sentence, but also a given sentence carries information on the context to which it belongs. When German native speakers see the sentence 3a/4b, for instance, they have a strong feeling about the context in which it occurs. It is very likely that the NP *ihren Mann* is stressed. It is either rhematic, or it carries contrastive focus. 1e is even more restricted. The personal pronoun *ich* must be contrastively stressed (I *myself* am the person who visits him). In every context requiring another stress, 1e is ungrammatical. It is thus possible to extract information on the context of a given sentence without having access to the preceding sentences.

Analysis grammars must allow most constituent order variations, as the number of phrase orders that can be excluded is very limited. The difference with generation grammars is that it is sufficient to generate one 'good' phrase order for each context, whereas in analysis all possible variations have to be allowed.

For this reason, the CF is of no use for analysis. Instead, analysis grammars should allow all grammatical orders and identify thematic, rhematic and focussed phrases.

In our algorithm, the number of possible themes and rhemes is limited to one constituent each, as this is sufficient to generate the variations in 5 to 7. Firstly, focus should be identified, and after this theme and rheme. Some permutations are only possible if one constituent is stressed contrastively. These constructions include the *Vorfeld* position of some typically rhematic elements (8, 9), the right movement of constituents which have a strong tendency to the left (cf. 1e and 5d above), and others (Steinberger 1994).

8 Nach FRANKreich$_{+focus}$ ist Vahé geflogen.
*To France is Vahé flew (Vahé flew to France)*

9 Einen INder$_{+focus}$ hat Anne geheiratet.
*An Indian has Anne married (Anne has married an Indian)*

In the next step, the theme category is identified. Every element at the beginning of the clause is marked as a theme if it has not been identified as a focus in the preceding step (10, 11):

10 Damals$_{+theme}$ lebte Hendrix noch.
*Then lived Hendrix still (Hendrix was still alive then)*

11 Ich glaube, daß Tina$_{+theme}$ oft kocht.
*I believe that Tina often cooks*

Similar to Hajičová et al.'s (1993) suggestion for English, and to Matsubara et al.'s (1993) for Japanese, the last constituent of the sentence will be recognised as rhematic, as rhemes tend to occur sentence-finally (cf. 5a and 6b). Our approach differs from Hajičová et al.'s, however, in that we prohibit some elements from being rhematic. In German, these inherently non-rhematic elements include personal pronouns, as well as a limited set of modifiers such as *wohl* in 12. Although some modifier groups tend to be potential rhemes, and others do not, most modifiers must be coded individually in the dictionary (Steinberger, 1994). Note that if inherently non-rhematic elements occur sentence-finally, it is likely that either the verb in V2 position, or the Vorfeld element carry heavy stress (12a vs. 12b).

12a Er LAS$_{+focus}$ den Artikel über Wortstellung dann *wohl*$_{-rheme}$ .

*He read the article on word-order then presumably*

12b ?? Er las den ArTIkel über Wortstellung dann *wohl*$_{-rheme}$ .

Hajičová et al. (1993) suggest that verbs are generally marked as rhemes, except if they have very general lexical meaning (such as *be, have, happen, carry out, become*). As our main concern is word order, and German verb placement is restricted by rules which do not allow variation, our algorithm does not allow the recognition of verbs as rhemes. In 12, no constituent would be recognised as being rhematic.

Not all languages express theme, rheme and focus as distinctly by word order variation as German does. Either they rely on the context to find out which constituents (have to) carry stress, or they use other means such as clefting, pseudo-clefting, topicalisation, dislocation, voice, impersonal constructions, particles, and morphological as well as lexical means (Foley/Van Valin 1985). However, even in English, which is often referred to as a fixed word order language, information on theme and rheme can be extracted automatically (Hajičová et. al. 1993; Steinberger 1992a). To which degree this information is conveyed in other languages, and by which means, must be subject to a language pair-specific investigation. The extraction of information on theme, rheme and focus is more important when translating from one free phrase order language into another, than when translating into a fixed-word order language. However, there are independent reasons for recognising the sentence focus, namely the correlation between stress on the one hand, and scope of negation (Payne 1985) and of degree modifiers (Steinberger 1992b) on the other.

# 7  Ambiguity Resolution

Findings on natural, less natural and ungrammatical word order variations can also be used to improve sentence analysis with respect to some cases of ambiguity resolution. In the case of 13, *eher* can be recognised as denoting *earlier* (eher$_{26}$), as the homonymous adverb (eher$_5$, "rather") must not be negated. Furthermore, some cases of unlikely PP attachment can be nearly excluded. In 14, the PP expressing location (vor der Bank) is unlikely to be a sentence modifier, as this would result in contrastive focussing of the personal pronoun *ihn*. This can be seen in 15, where the PP cannot be an adjunct to the preceding NP, because the NP is realised as a pronoun. The PP in 14 is thus more likely to be an adjunct to the nominative NP der Mann (14a) than a sentence modifier (14b). The general principle is that focussing constructions are relatively unlikely to occur in written text, and therefore one should avoid the analysis involving focus when another analysis is possible. This is the case when the analysis of the PP as an adjunct results in a sentence without contrastive stress.

13a Er sollte nicht eher$_{26}$ kommen. (not earlier)

*He should not earlier come (He should not come earlier)*

13b * Er sollte nicht eher$_5$ kommen. (rather)

*He should not rather come*

14 Deshalb hat der Mann vor der Bank ihn gesehen.

*Therefore has the man in-front-of the bank him seen (Therefore the man in front of the bank has seen him)*

14a Deshalb hat der Mann vor der Bank ihn gesehen.

14b ? Deshalb hat der Mann vor der Bank IHN ignoriert.

15 ?? Deshalb hat er vor der Bank IHN gesehen.

*Therefore has he in-front-of the bank him seen*

# 8  Conclusion

The order of constituents in free phrase order languages is determined by a set of factors which constitute tendencies rather than clear-cut rules. The fact that most, but not all, constituent orders are possible, and that some orders are more natural than others poses a considerable problem for NLP.

In this paper, we presented a method to deal with these problems from the analysis and the generation point of view. Concerning analysis, the main idea is that single sentences reflect the theme-rheme structure imposed by the context, so that thematic, rhematic and (contrastively) focussed constituents can often be recognised. In generation, we can convey this knowledge by differing word order depending on the context. This is achieved by using a canonical form which includes the *flexible* categories *theme*, *rheme* and *contrastive focus*.

A major advantage over methods suggested in the past is that acceptability differences between sentences can be dealt with, and that even modifier sequences, which are traditionally left out in word order description, can be handled. Wrong constituent orders are avoided, because the order of the major part of the sentence is fixed, and only single constituents move to the theme and rheme positions. The difficulty arising from the unclear borderline between free and fixed phrase order, which is typical of most *free phrase order languages*, is dealt with successfully.